\documentclass[10pt,twocolumn,letterpaper]{article}

\usepackage[pagenumbers]{cvpr}

\usepackage{preamble}

\definecolor{cvprblue}{rgb}{0.21,0.49,0.74}
\usepackage[pagebackref,breaklinks,colorlinks,allcolors=cvprblue]{hyperref}

\begin{document}

\title{\textit{DiT-Garment:} Garment Dynamics with Diffusion Transformers}

\author{Antoine Dumoulin \\
        Inria Centre at the University Grenoble Alpes\\
        {\tt\small antoine.dumoulin@inria.fr}
        \and
        Laurence Boissieux\\
        Inria Centre at the University Grenoble Alpes\\
        {\tt\small laurence.boissieux@inria.fr}
        \and
        Joao Regateiro\\
        InterDigital Inc.\\
        {\tt\small joao.regateiro@interdigital.com}
        \and
        Pierre Hellier\\
        Inria, University of Rennes, CNRS, IRISA-UMR 6074\\
        {\tt\small pierre.hellier@inria.fr}
        \and
        Stefanie Wuhrer\\
        Inria Centre at the University Grenoble Alpes \\
        {\tt\small stefanie.wuhrer@inria.fr}
}
\maketitle


\begin{abstract}
We present \our to model dynamic 3D clothing over human body models in arbitrary motion.
Unlike existing methods, \our can animate garments with unseen designs and physical materials, while allowing for direct inference of deformations for any target pose.
To achieve this, we leverage a 2D diffusion transformer architecture to learn 3D deformations in a 2D \uv-space.
As the result is non-deterministic, our generative model learns the distribution of possible outcomes.
The template garment is represented as a 3D triangle mesh spatially aligned with a 3D human body model in a standardized pose.
To work with different garment designs without the need of a common template or complex graph convolution operations, the diffusion transformer is conditioned on a 3D position map of the template, represented in \uv-space, which allows to implicitly learn a deformation of the 3D space around the body in standard pose. 
Further conditioning on body motion and physical parameters allows to physically ground the model. 
We quantitatively and qualitatively evaluate \our on both synthetic and real data. While only trained on synthetic simulations of automatically generated cloth designs, our method generalizes to captured and artist-made garment designs. Code and data are available for research purposes at \href{https://dumoulina.github.io/dit-garment/}{https://dumoulina.github.io/dit-garment/}.
\end{abstract}

\section{Introduction}
\label{sec:intro}

\begin{figure}
    \centering 
    \includegraphics[width=\linewidth]{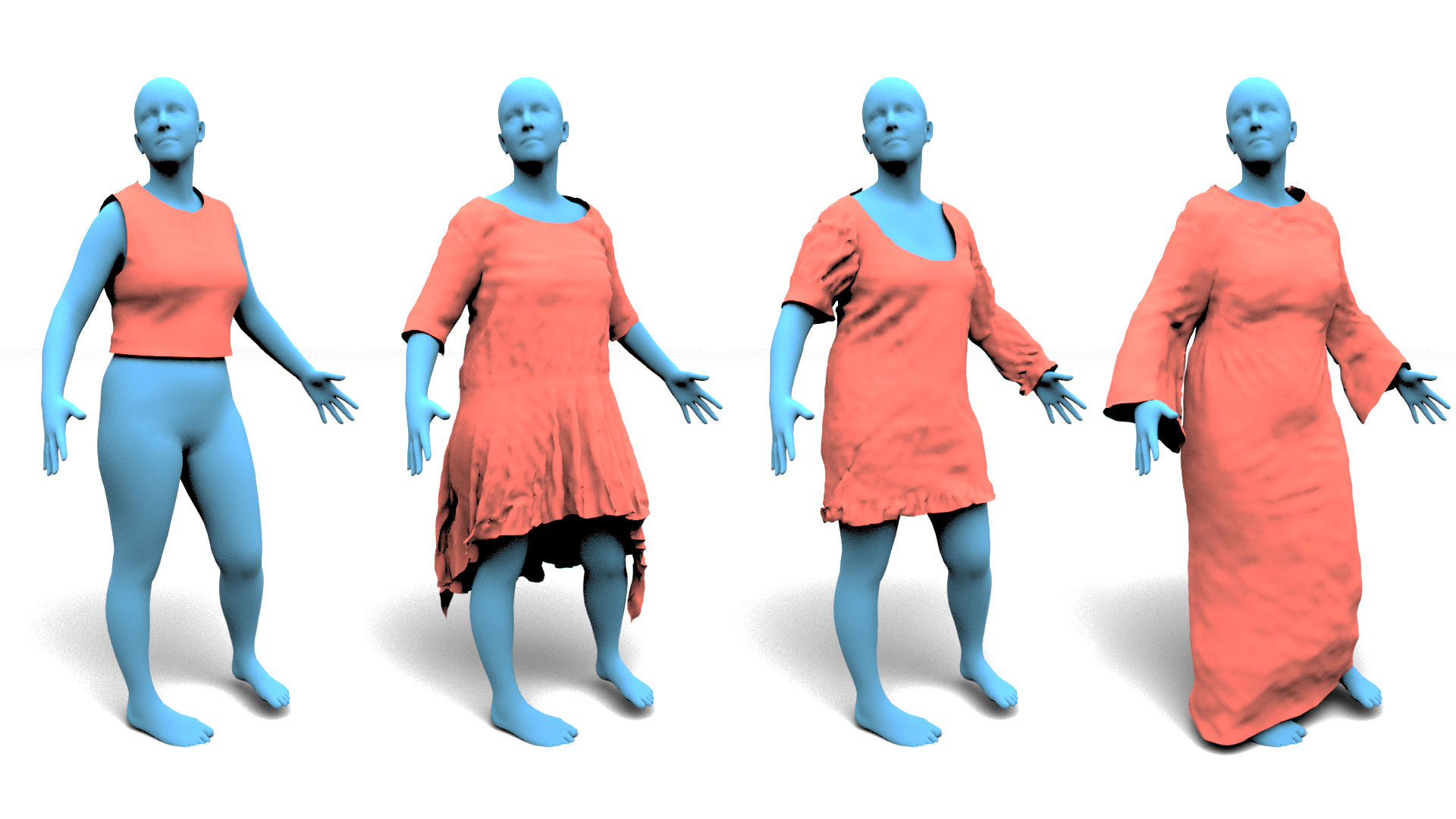}
    \caption{
        \our animates arbitrary garment designs over any human body in any target pose. Given a body motion and physical parameters, \our directly infers dynamic garment deformations for a design seen during training (left) and generalizes to three unseen designs (right), without retraining.
    }
    \label{fig:teaser}
\end{figure}

Garment modeling is a highly studied topic in computer vision and computer graphics. Dressing 3D avatars is a subject of interest in tele-presence applications such as augmented or virtual reality. Furthermore, in applications such as garment prototyping and virtual try-on, an important use case is to dynamically adapt garment geometry to various body morphologies and poses. In this context, modeling the dynamics of digital garments is an important problem.

We aim to simulate garments over human body models in arbitrary motion, with high control. In particular, it is desirable for users to control the garment design, the physical cloth material, and the wearer's body shape and motion.  Our objective is to animate a given input garment by deforming it to fit the last frame of the input body motion.

The problem of simulating garments over body models is well studied. Most works represent the garment using representations that are graph-based, point cloud-based, or use a 2-dimensional \uv-based mapping of the 3D garment. Graph-based methods~\cite{baraffsimulation, hood} are highly realistic and generally model dynamics well. However, most of these methods are auto-regressive, \ie they start from a standard pose and fit a full motion sequence up to the desired pose, rendering the methods complex for posing tasks. Existing methods can generalize to garment designs unseen during training, but lack the mesh connectivity or surface structure needed to model physical constraints such as stretch and bending strain, limiting their physical fidelity~\cite{pointbased, prokudin2023dynamic}. Methods that use a \uv-based representation~\cite{deepwrinkles, diffusedwrinkles} offer the advantage of directly benefiting from standard 2D neural architectures, and have recently been used to allow for fine-grained user control in case of a fixed garment design~\cite{dumoulin2026dgarment}. However, there is currently no method that allows for input garment designs unseen during training and physical material information, while directly inferring the deformations for the target pose and modeling garment dynamics.

To address this problem, we present \our, a physically grounded conditional generative model of garment deformation learned using diffusion transformers. We represent a garment design as a 3D template mesh, and the physical cloth material using three physical quantities describing bending, stretching and area density. Body motion is represented as a discrete sequence of parametric body poses along with a body shape. To work on a common space for all garment designs, we take advantage of a diffusion transformer model~\citep{sana} in a two-dimensional \uv-space to conditionally deform a given garment template.
\our provides fine-grained control over input conditions related to the person wearing the garment and the garment's design and material. Since \our only depends on a small sequence of body poses leading up to the desired pose, the result is not deterministic. To represent the diverse outcomes of garment deformations under unseen previous states of the garment, we leverage a generative model which learns a distribution of plausible garment deformations conditioned on the input conditions.

The core of our method is a garment representation that generalizes across designs unseen during training. We represent each garment as a 3D triangle mesh aligned to a human body model in a standard pose. Because all templates share the same body alignment, the model can learn corresponding deformations across different garment designs without explicit registration. The 3D mesh is then projected into 2D \uv-space as a position map encoding the 3D rest coordinates of each garment vertex. We use this position map as positional encoding of the diffusion transformer, so that attention over \uv coordinates conditions on rest-shape geometry rather than \uv layout. This design preserves seam connectivity: corresponding seam vertices map to shared 3D coordinates, and the attention mechanism learns smooth spatial correlations across \uv seams.

To train dynamic clothing models, we introduce a dataset composed of dynamically simulated clothes paired with human body motions. Existing synthetic cloth datasets such as Cloth3D~\citep{cloth3d} provide large-scale simulation data, but rely on a limited set of template designs. While resizing, cutting and reshaping the templates allows to generate a large number of garments, the underlying topology remains fixed.
Recent advances in garment modeling~\citep{GarmentCode} leverage sewing patterns to generate diverse and complex garment designs. To build our dataset of dynamic garments with diversified designs, we pick templates from GarmentCodeData~\citep{GarmentCodeData}. We then simulate the templates over various body shapes and motions with different physical materials.
This dataset is used to train our model and evaluate its generalization to unseen complex garment designs.
We further evaluate \our on real captured data from 4D-Dress~\citep{4ddress} and 4DHumanOutfit~\citep{4dhumanoutfit}.
Quantitative and qualitative results show that \our generalizes to unseen garment designs and physical materials, while producing realistic deformations at interactive frame rates.

In summary, our contributions are: 
\begin{itemize}
    \item A positional encoding that maps geometry into \uv-space, allowing 2D neural architectures to work on arbitrary mesh templates with varying \uv parametrizations.
    \item \our, a conditional generative model based on a \uv-based representation: We propose a diffusion transformer that enables direct posing of dynamic garments. Our model produces realistic deformations while providing fine-grained user control on physical properties, body shape and motion.
    \item An intersection-aware guidance strategy during the reverse diffusion process that effectively penalizes and mitigates cloth-body intersections for complex, unseen poses.
    \item A sewing-pattern-based dynamic dataset: We provide a comprehensive synthetic dataset featuring complex garment designs (from tight tanks to loose dresses) simulated over diverse body motions and physical materials. This dataset will be publicly released to foster further research.
\end{itemize}

\section{Related work}
\label{sec:relatedwork}

We organize existing works according to how the garment is represented. Common representations are graph-based, point cloud based, and 2-dimensional mappings of the 3-dimensional garments. A parallel line of work using implicit functions to model clothing~\cite{dig, smplicit, drapenet} will not be discussed as these works focus on shape modeling rather than cloth dynamics. 

\subsection{Graph-based representation}

Cloth is traditionally simulated using graph operations over 3D surface meshes~\citep{terzopoulos1987elastically, argus}.
Common physics-based approaches model cloth motion with particles~\citep{breen1994particles}, mass-springs~\citep{massspring} or finite elements~\citep{terzopoulos1987elastically}.
\citet{baraffsimulation} proposed a robust implicit solver for cloth simulation offering a good trade-off between accuracy and speed.
Faster alternatives have been proposed such as Projective Dynamics~\citep{projectivedynamics, simulator}, Position Based Dynamics \citep{pbd, xpbd} and Vertex Block Descent~\citep{chen2024vertex}.
However, time integration methods remain computationally expensive and require expertise to tune the physical parameters to achieve stable results~\citep{wang2011data}.

Recently, learning based methods have been proposed to render cloth simulation more accessible. TailorNet~\cite{tailornet} learns to predict static deformations of garments given a body pose.
To model dynamic deformations, template-specific methods~\citep{santesteban2021self, pbns, snug, gaps} learn to predict cloth dynamics over body motion.
Some existing works generalize to multiple garment designs in a single model.
GarSim~\citep{garsim} and HOOD~\citep{hood} learn to predict cloth dynamics for multiple garment designs leveraging a graph neural network. ContourCraft~\citep{contourcraft} propose a new contour loss to improve the quality of the generated garments with HOOD making it a strong baseline.
Recently, transformer-based models \citep{shi2024, li2024neural} show promising results in modeling cloth dynamics for multiple garment designs.

Most works relying on a graph-based representation are autoregressive starting from a standard pose, and cannot directly infer deformations for an arbitrary target pose. In contrast, \our allows for direct posing.

\subsection{Point cloud representation}

Point-based representations provide a flexible alternative to meshes for modeling clothed humans, as they do not impose explicit topological constraints on garment deformations. Early works \cite{scale, pointbased} explored point-based representations for modeling pose-dependent clothed human surfaces. Subsequent methods \cite{pop, qianli2022} addressed artifacts appearing with highly deformed regions of loose garments, where sparse point sampling can lead to incomplete or noisy surface reconstructions. FiTe \cite{lin2022fite} further introduced an implicit representation of the garment surface in a canonical pose, which is then explicitly deformed to the target pose.
Some point-based methods additionally use a hybrid \uv representation for the body geometry input \cite{scale, pop}.

Dynamic Point Fields (DPF) \cite{prokudin2023dynamic} instead learn deformation fields over an explicit source garment, using geometric constraints such as as-isometric-as-possible regularization.
The method can animate a 3D garment using SMPL vertices as a driving signal while maintaining correspondence with the source point representation. This ability to deform an existing garment template while preserving its geometric structure makes DPF a particularly relevant baseline for point-based garment deformation.

More recently, ClothDiffuse \cite{temporaldiffusion} introduced a diffusion-based framework for generating temporally coherent cloth deformations conditioned on body motion, enabling the modeling of dynamic garment motion. However, point-based representations primarily capture the garment surface geometry and its deformation, without explicit information describing the physical structure of the garment. This limits their ability to model physically grounded cloth dynamics, where quantities such as local connectivity, material properties, and interactions between neighboring surface elements are important. In contrast, \our explicitly models physics-inspired dynamics.

\begin{table}
    \centering
    \small
    \caption{
        Positioning of our method compared to existing learned-based approaches.
        \textbf{Gar.} indicates whether the method can generalize to garments unseen during training.
        \textbf{Motion} indicates whether the method models physical dynamics given body motion.
        \textbf{Phys.} allows for control of the physical material.
        \textbf{Posing} indicates whether the method can directly infer deformations for any target pose.
        \textbf{Bold} indicates compared methods in experiments.
    }
    \label{tab:positioning}

    \begin{tabular}{@{}l @{} c c c c @{}}
    Method & Gar. & Motion & Phys. & Posing \\

    \hline
    \hline
    \multicolumn{5}{l}{Graph representation} \\
    \hline
    TailorNet \cite{tailornet}               & \xmark & \xmark & \xmark & \cmark  \\
    \citet{santesteban2021self}              & \xmark & \cmark & \xmark & \xmark  \\
    PBNS~\citep{pbns}                        & \xmark & \cmark & \cmark & \xmark  \\
    SNUG~\citep{snug}                        & \xmark & \cmark & \xmark & \xmark  \\
    GAPS~\citep{gaps}                        & \xmark & \cmark & \xmark & \xmark  \\
    Cape~\citep{cape}                        & \cmark & \xmark & \xmark & \cmark  \\
    GarSim~\citep{garsim}                    & \cmark & \cmark & \cmark & \xmark  \\
    HOOD~\citep{hood}                        & \cmark & \cmark & \cmark & \xmark  \\
    \citet{shi2024}                          & \cmark & \cmark & \xmark & \cmark  \\
    \citet{li2024neural}                     & \cmark & \cmark & \xmark & \xmark  \\
    \textbf{ContourCraft}~\citep{contourcraft}& \cmark & \cmark & \cmark & \xmark  \\

    \hline
    \hline
    \multicolumn{5}{l}{Point cloud representation} \\
    \hline
    SCALE \cite{scale}                       & \xmark & \xmark & \xmark & \cmark  \\
    \citet{pointbased}                       & \cmark & \xmark & \xmark & \cmark  \\
    PoP \citep{pop}                          & \cmark & \xmark & \xmark & \cmark  \\
    FITE \citep{lin2022fite}                 & \cmark & \xmark & \xmark & \cmark  \\
    SkiRT~\citep{qianli2022}                 & \cmark & \xmark & \xmark & \cmark  \\
    ClothDiffuse \cite{temporaldiffusion}    & \cmark & \cmark & \xmark & \cmark  \\
    \textbf{DPF} \citep{prokudin2023dynamic} & \cmark & \xmark & \xmark & \cmark  \\

    \hline \hline
    \multicolumn{5}{l}{\uv-based representation} \\
    \hline
    DeepWrinkle \citep{deepwrinkles}          & \xmark & \xmark & \xmark & \cmark  \\
    \citet{motionguided}                     & \xmark & \cmark & \xmark & \xmark  \\
    D-Garment \citep{dumoulin2026dgarment}   & \xmark & \cmark & \cmark & \cmark  \\
    \citet{gan}                              & \cmark & \xmark & \xmark & \cmark  \\
    DiffusedWrinkles~\citep{diffusedwrinkles}& \cmark & \xmark & \xmark & \cmark  \\
    \textbf{Pyramid-Drape} \citep{pyramid}   & \cmark & \xmark & \xmark & \cmark  \\

    \hline
    \hline
    \our                                     & \cmark & \cmark & \cmark & \cmark  \\

    \end{tabular}
\end{table}

\subsection{\uv-based representation}

To leverage efficient neural network architectures in the 2D domain~\cite{unet, vit}, recent approaches encode 3D garment deformations in a 2D \uv intermediate representations. DeepWrinkles \cite{deepwrinkles} model static pose-dependent wrinkles using normal maps.
\citet{gan} propose a generative adversarial network to generate garments over body shape and pose. DiffusedWrinkles \cite{diffusedwrinkles} introduce a diffusion model to generate pose-dependent wrinkles over multiple garment designs.
Pyramid-Drape \cite{pyramid} propose a multi-scale architecture to generate pose-dependent wrinkles over multiple garment designs.
In addition to pose-dependent effects, \cite{motionguided, dumoulin2026dgarment} aim to model motion-dependent deformations.

\uv-based methods are particularly promising for garment modeling as they allow to benefit from standard 2D neural architectures. However, there is currently no method that allows simultaneously for generalization to garments unseen during training, allows for control of physical materials, models dynamics, and allows for posing. \our is a \uv-based method that unlike previous work learns to deform arbitrary mesh templates with varying topology and \uv parametrization.

\begin{figure*}
    \centering 
    \includegraphics[width=\linewidth]{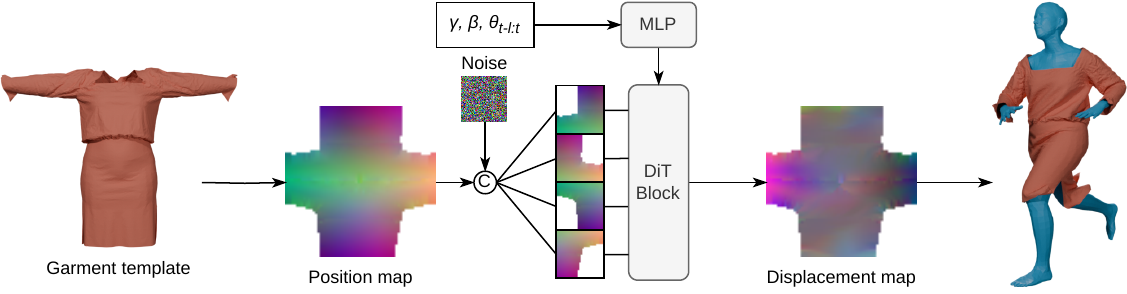}
    \caption{
    \our generates garment deformations for a given template representing a garment design conditioned on body shape~\bodyshape, motion~\bodymotion and cloth material~\material (see \cref{sec:representation}). It builds upon a 2D diffusion transformer model (DiT) to learn how to deform a template in \uv-space. 3D geometric features are parameterized by the \uv parametrization of the template, and we concatenate the position map to the diffusion noise. At inference, our model deforms the garment by iteratively denoising the Gaussian noise to the displacement map.
    }
    \label{fig:model}
\end{figure*}

\subsection{Positioning}

Table~\ref{tab:positioning} summarizes the positioning of \our \wrt existing works. \our is the first \uv-based method that simultaneously allows to model unseen garments during training, to control the physical cloth material, to simulate dynamics over body motion, and to infer cloth deformations from any pose. We experimentally compare \our to strong baselines from each representation category: graph-based, point-based and \uv-based methods.

\section{Method}
\label{sec:method}

We introduce \our, a unified model that learns to generate dynamic cloth deformations to dress 3D avatars in motion for multiple garment designs, represented by templates. \Cref{fig:model} illustrates the architecture, which is based on a Diffusion Transformer (DiT)~\citep{diffusiontransformers} operating in \uv-space. By aligning all garment templates to a common rest pose and projecting them into a two-dimensional parameterized space, \our bypasses the need for shared topology, explicit rigging, or complex graph operations. Instead, \our learns to predict three-dimensional deformations by leveraging the spatial layout encoded in the template's position map.

Given a mesh template representing a garment, the wearer's body shape, a sequence of body poses, and the cloth's physical material parameters (\cref{sec:representation}), our approach deforms the mesh template to fit the body in the last frame of the sequence of body poses. This is achieved by representing the garment geometry in a 2-dimensional \uv space, and by learning how to displace each vertex of the template. The body and physical material are represented using low-dimensional vectors as in D-Garment~\citep{dumoulin2026dgarment}.

The model operates in a non-autoregressive manner: at each frame, it predicts the target deformation directly from the current input, without requiring current cloth velocity, or deformation history. Because no predicted state feeds back into subsequent frames, this frame-independent generation avoids the error accumulation typical of autoregressive simulators. Temporal coherence across a motion sequence arises from the smooth variation of the driving input rather than from any temporal recurrence in the model.

It is important that the mesh templates of all garments are aligned in $\mathbb{R}^3$, as we use this alignment to encode correspondence information between different garments. In practice, we align all templates on a human body model in fixed body shape and pose. The unposed template is then unwrapped to a 2D position map which encodes the rest shape geometry of the garment design in a 2D image. The position map acts as a structural guide, informing the DiT about the template's layout and ensuring that generated displacement maps align with the input's \uv coordinates.
Because the position map encodes the 3D spatial arrangement of vertices, the model can learn correspondence information across different garment templates without requiring explicit \uv-map correspondence. Consequently, a single trained model generalizes to multiple garment designs with varying \uv parametrizations.

We train the model on a dataset of simulated cloth featuring diverse designs, motions, and physical parameters (\cref{sec:dataset}). At inference, \our enables dynamic simulation (\cref{sec:exp-simulation}) of unseen garment designs.

\subsection{Garment modeling and representation}
\label{sec:representation}

We aim to geometrically deform a 3D model of a garment template \mesh, represented as triangle mesh, to $\mesh_t$ at a discrete time step $t$.
The dynamic deformations of the garment \mesh to $\mesh_t$ are modeled as the offset vector $v_t \in \mathbb{R}^{|V|\times3}$ where $V \in \mathbb{R}^{n\times3}$ is the set of $n$ vertices of \mesh.
To learn a normalized space of garment templates, we align all templates to a standard rest pose in $\mathbb{R}^3$ where the garment is unposed and aligned with a human body model in a fixed pose. This alignment enables the model to learn correspondences across different garment designs.

Inspired by \textit{geometry images} \citep{gu2002geometry} and their application in cloth deformation models \citep{pyramid, dumoulin2026dgarment}, we map 3D information onto 2D images. Unifying garment mesh templates with varying connectivity and vertex resolution to a standard 2D grid allows us to use state-of-the-art 2D architectures.
Unlike previous work, our method directly learns from arbitrary mesh templates with varying \uv parametrization.

We denote the mapping function projecting 3D mesh surface to image pixels by $\phi \colon \mathbb{R}^3 \mapsto \mathbb{R}^2$, and its inverse applying 2D mapping to mesh vertices by $\phi^{-1} \colon \mathbb{R}^2 \mapsto \mathbb{R}^3$.
We use $\phi$ to encode geometric features including position map $\phi(\mesh)$ and displacement map $\phi(\mesh_t)$.
Thus, \our learns to generate the displacement map $\phi(\mesh_t)$ from the position map $\phi(\mesh)$ and the other conditional inputs.
Note that each garment \mesh has its own \uv parametrization $\phi$.

\our represents a dynamic garment on top of a parametric human body model that decouples body shape and pose parameters. By representing all garment templates in the same standard pose, \our learns the relationship between the garment and the body without requiring explicit rigging or correspondence information between the different garments. In our implementation, we use SMPL~\citep{smpl}, and represent body shape \bodyshape and pose sequence \bodymotion as concatenation of the $l$ preceding poses and the current one at time $t$. \bodypose includes both the global and local body joint rotations and positions. \bodyshape is represented as a low-dimensional vector of shape coefficients.

To physically ground \our, the network uses cloth material information. The representation of cloth material is inspired by physics-based cloth simulation~\citep{baraffsimulation} and includes stretch coefficient \stretching (in $N/m$), mass density coefficient \density (in $kg/m^{2}$) and bending coefficient \bending (in $N \cdot m$) as $\material := [\stretching,\density,\bending]$. Parameter \stretching controls resistance to stretching or compression, \density controls the influence of inertia, and \bending controls resistance to bending or curvature changes. 

\our, shown in \Cref{fig:model}, can be formulated as:
\begin{equation}
 \mesh_t \sim \mathcal{G}(\condition).
\end{equation}

\subsection{Diffusion transformer model}
\label{sec:diffusion}

The 3D mesh generator $\mathcal{G}$ is built on a 2D diffusion transformer (DiT)~\citep{diffusiontransformers}, leveraging the recent success of diffusion models for 3D cloth deformation in \uv-space~\citep{dumoulin2026dgarment, diffusedwrinkles, diffusionprior, guo2025high, manipulated, garec}. Unlike prior works, the transformer architecture can learn from diverse garment templates and \uv parametrizations. This is thanks to the position map $\phi(\mesh)$ that encodes the 3D coordinates of the garment vertices in \uv-space, which allows to implicitly learn a deformation of the 3D space around the body in standard pose. The position map is concatenated to the diffusion noise and fed to the DiT, giving the necessary information to generate displacement maps aligned with the input \uv parametrization. The model is further conditioned on body shape, body motion, and cloth material to produce physically grounded deformations. The output of the model is a displacement map $\phi(\mesh_t)$ that encodes the deformation of the garment from its rest pose to the target pose.

During training, the conditional denoising transformer $\epsilon_\theta$ is trained with rectified flow objective~\citep{rectifiedflow} using samples $\{\mathcal{M}_t,\condition\}$ from our training data corpus as:
\begin{equation}
\mathbb{E}_{\mathbf{z}, \epsilon, s} \left[ \|(\epsilon-\mathbf{z}_0) - \epsilon_\theta(\mathbf{z}_s, s, \condition) \|_2^2 \right],
\end{equation}
where $\mathbf{z}_0 = \phi(\mesh_t)$, $\epsilon \sim \mathcal{N}(0, I)$, $s$ the diffusion time step, and the noisy latent obtained with forward diffusion $\mathbf{z}_s = \alpha_s \mathbf{z} + \sigma_s \epsilon$. The scaling factor and standard deviation of the forward diffusion are
\begin{equation}
\bar{\alpha}_s = \prod_{u=1}^{s} (1 - \beta_u),
\quad \alpha_s = \sqrt{\bar{\alpha}_s},
\quad \sigma_s = \sqrt{1 - \bar{\alpha}_s}, 
\label{eq:scale}
\end{equation}
where variance $\beta_u$ is determined by the noise schedule. 

At inference, we sample the garment deformation ${\mesh}_t$ from the learned distribution by iteratively denoising a Gaussian noise $\mathbf{z}_S \sim \mathcal{N}(0, I)$ to the predicted displacement map ${\mathbf{z}}_0$ using the reverse diffusion process.

\subsection{Intersection guidance}
\label{sec:guidance}

While the trained model can generate realistic cloth deformations, it may produce intersections with the body. To mitigate this issue, we introduce a guidance strategy that penalizes cloth-body intersections during diffusion.

Inspired by~ \cite{jinlongloss}, $\mathcal{L}_c$ computes the distance of points inside the body to the closest point on the body surface:
\begin{equation}
\mathcal{L}_c = \sum_{p \sim \uniform(\mesh_t)} \delta_{\text{in}}(p,\mathcal{B})
\min_{b \sim \uniform(\mathcal{B})}||p-b||_2,
\label{eq:clothing-penetration-loss}
\end{equation}
where $p$ and $b$ are points uniformly sampled over the mesh surface, and $\delta_{\text{in}}$ is an indicator function for colliding points.

This loss guides the diffusion process by computing the gradient of $\mathcal{L}_c$ \wrt the generated displacement map $\phi(\mesh_t)$ and applying it as correction to the predicted flow of the denoiser output. Guidance is applied at each denoising step after a warmup period to remove initial noise.

\begin{figure*}
    \centering
    \includegraphics[width=\linewidth]{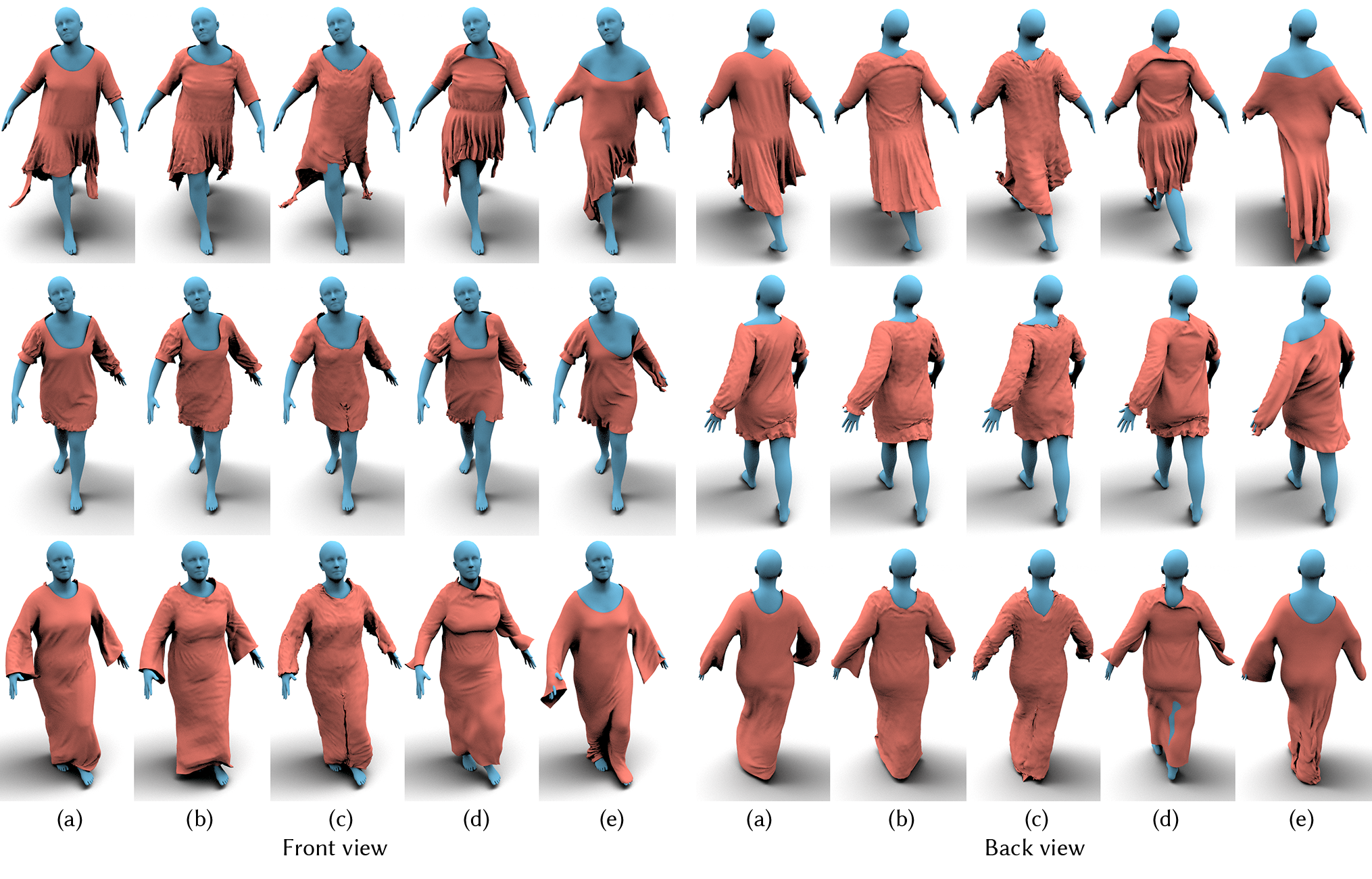}
    \caption{
    Visual comparison of the simulation ground truth (a), \our (b), Pyramid-Drape \cite{pyramid} (c), DPF~\cite{prokudin2023dynamic} (d) and ContourCraft \cite{contourcraft} (e). \our generates realistic deformations for unseen garment designs, physical materials and motions.
    }
    \label{fig:qualitative}
\end{figure*}

\section{Dataset}
\label{sec:dataset}

Existing datasets for cloth simulation, such as Cloth3D~\citep{cloth3d}, VTO~\citep{santesteban2021self} or D-Garment~\cite{dumoulin2026dgarment}, provide large-scale data but rely on a limited set of template designs. While resizing, cutting and reshaping these templates allows generating a large number of garments, the underlying topology remains fixed, limiting the diversity of garment designs. To address this limitation, we simulate a new dataset with diverse garment designs, physical materials, and body motions.

We build a garment template set of 22 upper clothes from GarmentCodeData~\citep{GarmentCodeData} with styles ranging from tight tanks to complex loose dresses. These templates are generated from sewing patterns based on GarmentCode~\citep{GarmentCode}, which allows for complex garment designs with varying structure, cut and sizing.
The templates are aligned to a human body model in a standard pose and unwrapped to a 2D \uv-space with Blender. ProjectiveFriction~\citep{simulator} simulates the garments over various body shapes and motions with different physical materials.
We select 169 body motions from AMASS~\citep{amass} that we discretize as pose sequences at 30 frames per second (FPS).
We augment the motion set with 3 random body shapes $\bodyshape \sim \mathcal{U}(-1, 1)^8$ per sequence. We then randomly sample 3 garments with materials in $\log(\bending) \sim \mathcal{U}(-8, -4)$, $\stretching \sim \mathcal{U}(40, 200)$ and $\density \sim \mathcal{U}(0.01, 0.7)$. In total, the dataset contains 1494 simulated sequences for a total duration of 2 hours.

For testing, 3 garment templates and 3 body motions are kept out from the training set.
The test set is particularly challenging, as it contains only sequences with unseen garment templates, unseen motions, and unseen body shapes. There are 27 sequences in the test set.

\begin{table}
    \centering
    \footnotesize
    \setlength{\tabcolsep}{3pt}
    \caption{
    Comparison on synthetic test set to Pyramid-Drape~\cite{pyramid}, DPF \cite{prokudin2023dynamic} and ContourCraft~\citep{contourcraft}. Best scores among methods without ground truth information are in \textbf{bold}, and - means that the method does not produce the output.
    }
    \label{tab:simulated-eval}

    \begin{tabular}{@{} l | *{3}{r} | *{4}{r} | r @{}}
     & \multicolumn{3}{c|}{\textbf{Shape Sim.}} & \multicolumn{4}{c}{\textbf{Phys. Validity}}\\
     & $E_v^\downarrow$ & $E_{CD}^\downarrow$ & $E_n^\downarrow$ & $E_c^\downarrow$ & $E_b^\downarrow$ & $E_s^\downarrow$ & $E_d^\downarrow$ & FPS \\
    \hline
    \hline
    \multicolumn{9}{l}{Method with access to ground truth correspondence and normals}\\
    \hline
    Pyramid-Drape$^*$ & - & 0.01 & 0.31 & 0.63 & - & - & 2.24 & 12.6 \\
    \hline
    \hline
    \multicolumn{9}{l}{Methods without ground truth information}\\
    \hline
    \our & \textbf{5.20} & \textbf{0.17} & \textbf{0.38} & 0.62 & 0.51 & 1.32 & \textbf{2.66} & 0.4 \\
    DPF  & 6.95 & 0.18 & 0.42 & 0.64 & \textbf{0.49} & \textbf{0.22} & 3.94 & 0.05 \\
    ContourCraft & 15.35 & 1.21 & 0.51 & \textbf{0.49} & 0.53 & 14.44 & 11.33 & 1.7 \\
    \end{tabular}
\end{table}

\section{Experiments}
\label{sec:experiments}

\subsection{Implementation details}

To learn the mapping from the input conditions to the garment deformation, we leverage a DiT architecture~\citep{sana} based on efficient Linear Attention. This model works without positional encoding which has been required by previous DiT models to infer a spatially coherent and structured image. The conditions {\bodyshape, \bodymotion, \material} are encoded with a 2-layer MLP and injected through cross-attention.
At inference, the guidance is applied after half of the 100 denoising steps with a guidance weight of 5.

Trained on 2 NVIDIA L40S for 100 epochs during 9 days, we use a batch size of 32 and a learning rate of $10^{-4}$ with Adam optimizer with weight decay of $10^{-2}$. The input \uv maps are $64\times64$ pixels. We sampled the training diffusion steps $s$ from a uniform distribution $\mathcal{U}(0, 1000)$ on a linear noise schedule with $\beta_0 = 10^{-4}$ and $\beta_{1000} = 0.02$.

\begin{figure}
\centering
\begin{tabular}{@{}cc}
    \includegraphics[width=0.5\linewidth]{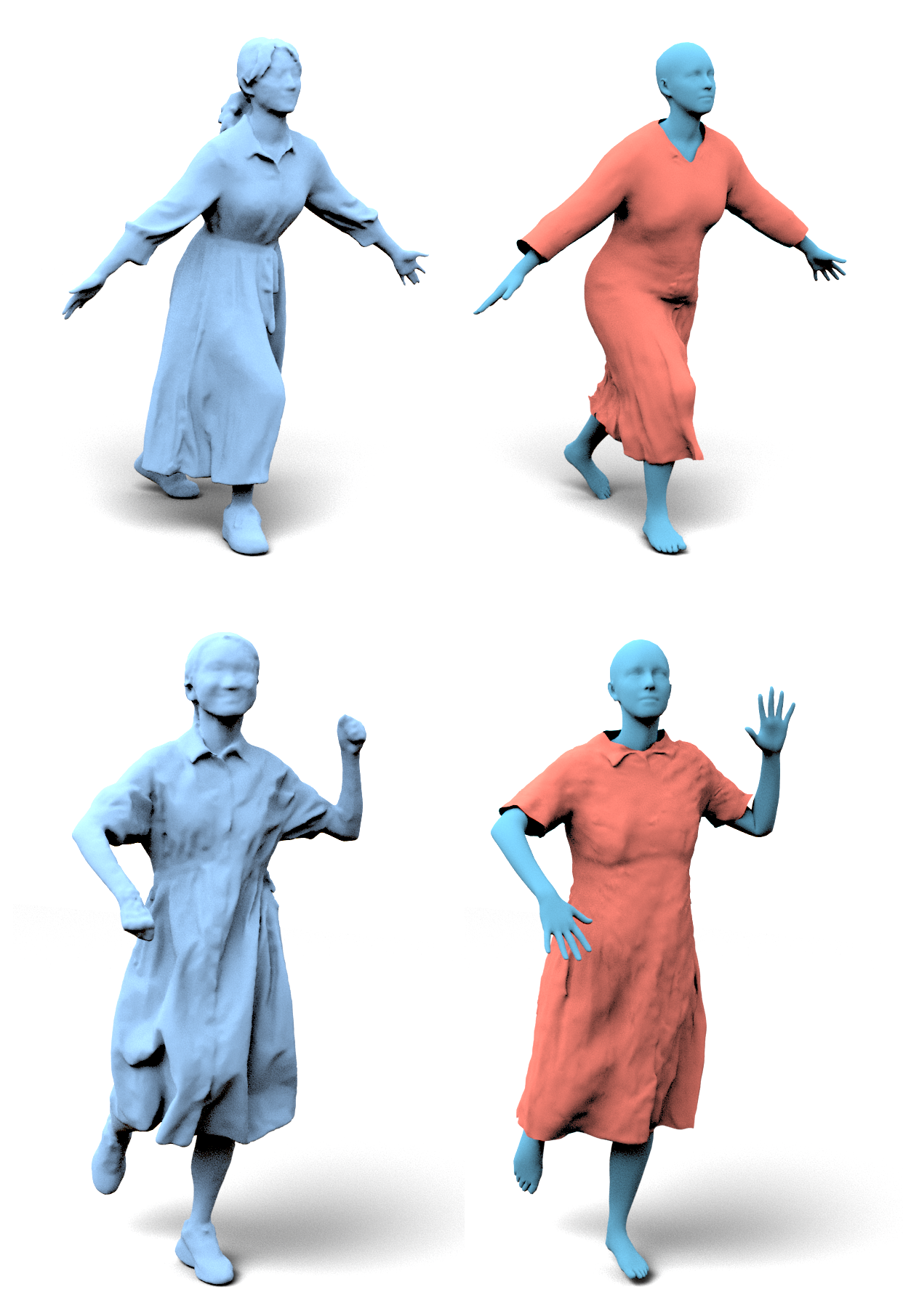}&\includegraphics[width=0.5\linewidth]{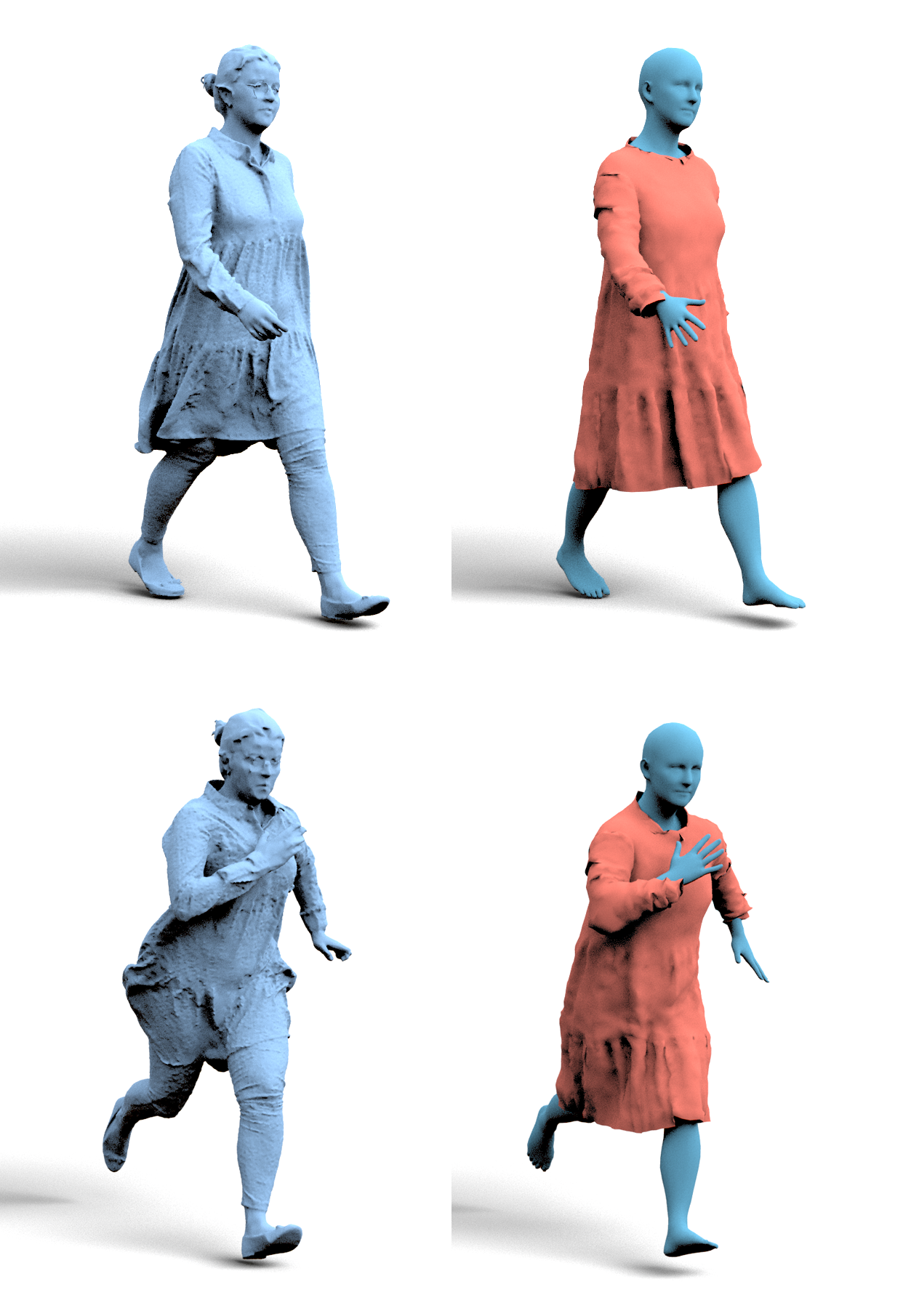}\\
    4D-Dress & 4DHumanOutfit
\end{tabular}
    \caption{
        We use \our to reproduce captured sequences from 4D-Dress~\citep{4ddress} and 4DHumanOutfit~\citep{4dhumanoutfit}.
        For each sequence, we show the captured avatar (left) and the predicted garment on top of the fitted body (right). We use reconstructed garment templates for 4D-Dress and an artist designed template for 4DHumanOutfit. 
    }
    \label{fig:real-data}
\end{figure}

\subsection{Evaluation measures}
\label{sec:metrics}

We follow previous work~\cite{dumoulin2026dgarment} and evaluate the results using both shape similarity and physical validity. 
$E_v$ is the average vertex-to-vertex distance between the predicted and ground truth meshes. $E_{CD}$ is the Chamfer distance comparing shape similarity. $E_n$ is the Chamfer normal distance comparing the wrinkling similarity.
$E_c$ is the percentage of cloth inside the body. $E_b$, $E_s$ and $E_d$ are the average bending, stretching and density errors relative to the ground truth, comparing the physical validity of the predicted garment.
$E_{CD}$, $E_v$ and $E_d$ measures are computed in centimeters.
We also indicate the inference speed in frames per second (FPS) for each method without pre-processing steps.

\begin{figure}
    \centering
    \includegraphics[width=0.7\linewidth]{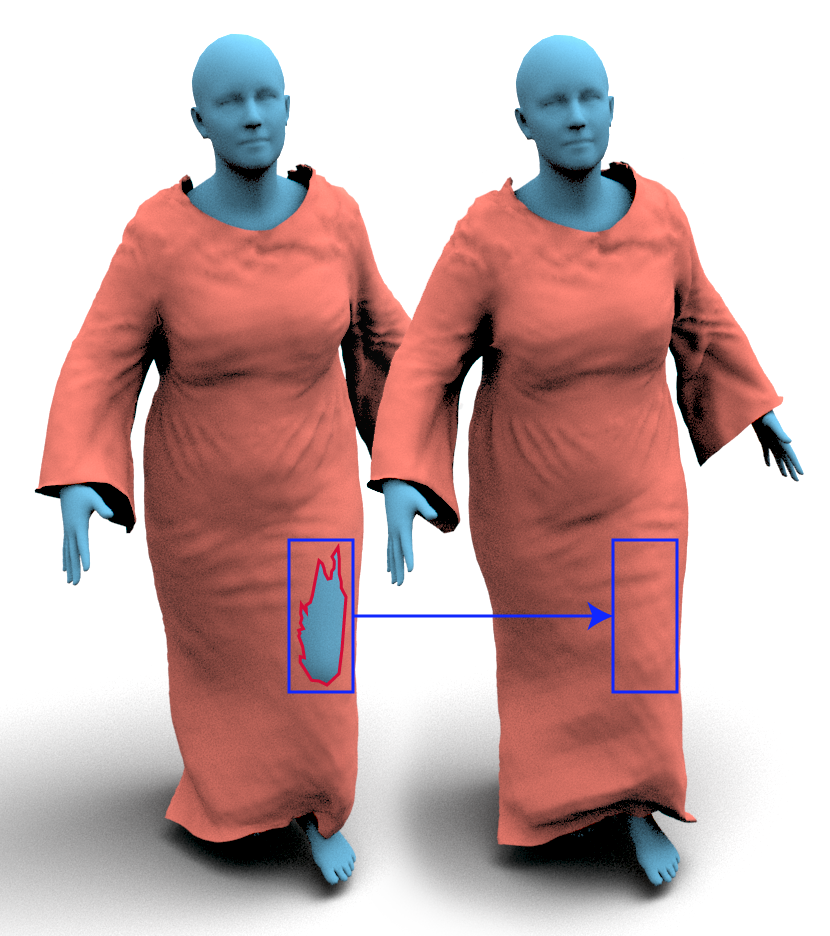}
    \caption{
        Effect of our guidance strategy to reduce cloth-body intersections. Left: without guidance, the generated garment intersects with the body. Right: with guidance, the generated garment does not intersect with the body.
    }
    \label{fig:guidance}
\end{figure}

\begin{table*}
    \centering
    \small
    \setlength{\tabcolsep}{5pt}

    \caption{
    Comparison of different transformer input patch size without guidance on seen and unseen templates. Smaller patch size improves the performance on unseen templates at the cost of inference speed. Guidance and patch size of 1 are used in the full model.
    }
    \label{tab:ablation}
    \begin{tabular}{@{} c | *{3}{r} | *{4}{r} | *{3}{r} | *{4}{r} | r @{}}
     & \multicolumn{7}{c|}{\textbf{Seen templates}} & \multicolumn{7}{c|}{\textbf{Unseen templates}}\\
    Patch & \multicolumn{3}{c|}{\textbf{Shape Similarity}} & \multicolumn{4}{c|}{\textbf{Physical Validity}} &  \multicolumn{3}{c|}{\textbf{Shape Similarity}} & \multicolumn{4}{c|}{\textbf{Physical Validity}} \\
     size & $E_v^\downarrow$ & $E_{CD}^\downarrow$ & $E_n^\downarrow$ & $E_c^\downarrow$ & $E_b^\downarrow$ & $E_s^\downarrow$ & $E_d^\downarrow$ & $E_v^\downarrow$ & $E_{CD}^\downarrow$ & $E_n^\downarrow$ & $E_c^\downarrow$ & $E_b^\downarrow$ & $E_s^\downarrow$ & \multicolumn{1}{r|}{$E_d^\downarrow$} & fps\\ 
    \cline{1-16}
8 & 3.07 & 0.10 & 0.28 & 0.65 & 0.48 & 0.15 & 1.44 & 5.99 & 0.25 & 0.44 & 0.64 & 0.51 & 1.17 & 2.89 & 11.9 \\
2 & 2.87 & 0.09 & 0.28 & 0.65 & 0.48 & 0.12 & 1.32 & 5.67 & 0.18 & 0.39 & 0.63 & 0.50 & 0.64 & 3.38 & 6.1 \\
1 & 3.09 & 0.09 & 0.28 & 0.64 & 0.48 & 0.12 & 1.31 & 5.38 & 0.20 & 0.38 & 0.62 & 0.50 & 0.62 & 2.99 & 2.6 \\
full & 3.02 & 0.09 & 0.28 & 0.64 & 0.48 & 0.13 & 1.33 & 5.20 & 0.17 & 0.38 & 0.62 & 0.51 & 1.32 & 2.66 & 0.4 \\
    \end{tabular}
\end{table*}

\subsection{Comparison on simulated data}
\label{sec:exp-simulation}

To evaluate the generalization of \our to unseen garment designs, we compare to methods that can animate any garment design. Pyramid-Drape~\citep{pyramid}, DPF~\citep{prokudin2023dynamic} and ContourCraft~\citep{contourcraft} are strong baselines for \uv-based, point cloud and graph-based representations, respectively. Pyramid-Drape and DPF do not allow for control of the physical material, while we inform \our and ContourCraft of the simulation physical material. We use pretrained models for Pyramid-Drape and ContourCraft when available, and optimize DPF on each body frame following the original implementation.

Figure~\ref{fig:qualitative} shows frames from test sequences of unseen garment designs along with predicted deformations from \our, Pyramid-Drape, DPF and ContourCraft.
While having access to ground truth information, Pyramid-Drape fails to generate wrinkle details and changes the garment topology. This leads to visual artifacts as in the bottom row (c). DPF generalizes well to complex garments, but lacks physical realism with cloth-body intersections and unrealistic wrinkles. ContourCraft fails to represent pleats and folds in the garment, producing oversmoothed deformations. In contrast, \our generates realistic deformations with detailed wrinkles and with few cloth-body intersections. We provide additional qualitative results in the supplementary video.

\Cref{tab:simulated-eval} reports the average shape similarity and physical validity metrics of all methods applied over the test set. Unlike all other methods, Pyramid-Drape has access to ground truth information as input. It changes mesh topology; hence, $E_v$, $E_b$ and $E_s$ cannot be measured on the output. Pyramid-Drape achieves the best $E_{CD}$ and $E_n$ errors.
Of the remaining methods, overall, \our presents competitive performance in most metrics, namely $E_v, E_{CD}, E_n$, and $E_d$. DPF achieves better $E_b$ and $E_s$ errors, thanks to its isometric loss.
ContourCraft achieves less cloth-body intersections $E_c$, minimized by its contour loss, but shows high $E_s$ for 10 fast sequences due to the slow propagation of graph neural networks, which leads to a high average error.

\our outperforms existing approaches without ground truth correspondence input in shape similarity while maintaining competitive physical validity metrics.

\subsection{Generalization on real data}

Previous experiments have shown that \our generalizes to unseen garment designs synthesized from a sewing pattern model \cite{GarmentCodeData}. To further evaluate the sim-to-real applicability of our model to real-world garments, we compare the generated deformations to captured garments from 4D-Dress~\citep{4ddress} and 4DHumanOutfit~\citep{4dhumanoutfit}. We run our model on the fitted body motion from the captured sequences, using a reconstructed garment template for 4D-Dress and an artist designed template for 4DHumanOutfit. The results are shown in \Cref{fig:real-data}. We use the same physical material parameters for all garments, which we arbitrarily set to $\stretching=50$, $\density=0.01$ and $\bending=10^{-8}$.
The animated garments generated by \our are visually plausible, showing that it can handle templates coming from real data while being only trained on synthetic data.
We provide additional qualitative results in the supplementary video.

\subsection{Ablations}

We evaluate the impact of our guidance strategy and the transformer patch size on the generalization capability of our model to unseen garment templates in \Cref{tab:ablation}. In this experiment, we compare the results with different patch sizes without guidance and with a patch size of 1 with guidance, which is used in the full model.
The patch size is one of the main hyperparameters of our transformer architecture: larger patch size implies smaller number of tokens and thus faster inference speed. However, larger patch size also implies that the attention has less information about the local structure of the garment template, which can affect the generalization capability of the model to unseen templates.
The quantitative results shows that smaller patch size improves the model performance on unseen templates. Notably, the patch size does not significantly affect the performance on seen templates showing that the model can learn to generate realistic deformations for seen templates even with a larger patch size. 
In practice, we use a patch size of 1 for all our experiments.
Our guidance approach is effectively removing remaining cloth-body intersections as shown in \Cref{fig:guidance} while maintaining the shape similarity and physical validity of the generated deformations in \Cref{tab:ablation}. Note that guidance requires additional diffusion steps to converge, which reduces the inference speed of the model. We only use 20 steps for testing our model without guidance compared to the 100 steps in the full model.

\subsection{Limitations}

\our is a data-driven model that learns to generate garment deformations from a dataset of simulated cloth. It is limited by the diversity and quality of the training data. The model may not generalize well to garment designs, body shapes, or motions that are significantly different from those seen during training.

In addition, our model does not explicitly enforce physical constraints, such as collision avoidance or energy conservation, which may lead to unrealistic deformations in some cases. While our guidance strategy mitigates cloth-body intersections, it does not guarantee collision-free results. Future work could explore strategies to incorporate more physical constraints into the model.

Finally, our model operates in a non-autoregressive manner, which has the advantage of avoiding error accumulation over time while still producing temporally coherent results, but may limit its ability to capture long-term temporal dependencies in garment dynamics.

\section{Conclusion}
\label{sec:conclusion}

We presented \our, a diffusion-based method for dynamically posing 3D garments conditioned on body shape, motion, and physical material properties. Our approach leverages a 2D diffusion transformer architecture operating in \uv-space, enabling generalization across diverse garment templates without requiring shared topology or explicit rigging. The position map serves as a structural guide, encoding 3D spatial relationships that allow the model to learn correspondences across garment designs.

To train our model, we introduced a new dataset of simulated cloth with various complex garment designs, physical materials, and body motions. 
Experiments demonstrate that \our produces realistic garment deformations at interactive frame rates, generalizing to unseen garments. 

\section{Acknowledgments}
This work was partially funded by the Nemo.AI laboratory by InterDigital and Inria.
Experiments presented in this paper were carried out using the Abaca infrastructure, supported by Inria (see https://abaca.inria.fr). We thank Adnane Boukhayma for helpful discussions.
We thank Hunor Laczkó for help with the comparison to Pyramid-Drape.

{
    \small
    \bibliographystyle{ieeenat_fullname}
    \bibliography{references}
}

\clearpage
\clearpage
\setcounter{page}{1}
\maketitlesupplementary

\section{Evaluation on unseen factors}

\Cref{tab:factors} evaluates \our performance on each input parameter individually. The unseen factors include garment design, body motion, body shape, and physical material.
Each test set contains the same 9 sequence input data sampled from our training set, resimulated with one specific factor replaced by an unseen value. We report the mean error for each metric as described in \Cref{sec:metrics}.
Results show that \our robustly generalizes to each unseen factor, with the lowest shape similarity errors for unseen physical material and the highest errors for unseen garment design. 
The lower $E_c$ observed for unseen body motions is primarily due to the specific body topology of the sampled test motions, which feature fewer complex self-intersections compared to the broader training distribution.
The higher errors in most metrics for unseen garment design can be explained by more complex designs than those in the training set, particularly one loose dress covering most of the body.

\begin{table}[h]
    \centering
    \small
    \setlength{\tabcolsep}{4pt}
    \caption{
    Evaluation of our method on one unseen factor at a time. Each test set uses the same sampled sequences from our training set, but with one factor replaced by an unseen value (garment design, body motion, body shape, or physical material).
    }
    \label{tab:factors}
    \begin{tabular}{@{} c | *{3}{r} | *{4}{r} @{}}
    Unseen & \multicolumn{3}{c|}{\textbf{Shape Similarity}} & \multicolumn{4}{c}{\textbf{Physical Validity}} \\
    factor & $E_v^\downarrow$ & $E_{CD}^\downarrow$ & $E_n^\downarrow$ & $E_c^\downarrow$ & $E_b^\downarrow$ & $E_s^\downarrow$ & $E_d^\downarrow$ \\ 
    \hline
garment design & 4.30 & 0.11 & 0.36 & 1.55 & 0.53 & 1.47 & 2.30 \\ 
body motion & 3.40 & 0.03 & 0.22 & 0.68 & 0.40 & 0.10 & 0.74 \\ 
body shape & 2.07 & 0.03 & 0.22 & 1.44 & 0.43 & 0.11 & 1.01 \\ 
phys. material & 0.94 & 0.01 & 0.19 & 1.68 & 0.43 & 0.11 & 0.85 \\
    \end{tabular}
\end{table}

\section{Statement of Broader Impact}

The datasets used in this paper contain motion captures of real people. 4D-Dress and 4DHumanOutfit followed rigorously ethics guidelines and GDPR rules. We used these datasets under permission given by their respective owners.

\paragraph{Positive Societal Impact.} DiT-Garment enables realistic, physically grounded animation of 3D garments on human avatars, which has broad applications in virtual try-on, digital fashion, gaming, film, and healthcare. By supporting unseen garment designs and other factors, our method lowers the barrier for creative industries to produce customized, physically realistic virtual clothing without expensive per-design simulation. The ability to generalize across captured, artist-made, and synthetic designs also promotes accessibility in virtual fashion, potentially reducing waste and promoting more sustainable practices in the apparel industry through more accurate digital prototypes.

\paragraph{Potential Societal Risks.} Realistically animating garments on human avatars in arbitrary poses raises privacy and misuse concerns. In particular, realistic virtual humans could be used to generate non-consensual or deceptive imagery, or to infer sensitive body-related information from public or private data. Additionally, because our model is trained on synthetic body models, there is a risk that it may not generalize equitably across diverse body shapes, potentially introducing biases in virtual fashion applications. The environmental cost of training and running large diffusion-based models should also be considered.

\paragraph{Mitigation and Responsible Use.} We commit to release code and data to encourage open, transparent research while encouraging downstream users to implement safeguards, such as access controls, watermarks, or consent mechanisms, when deploying models capable of generating realistic virtual humans. We also encourage evaluating fairness across diverse body types to reduce biases in 3D human avatars. Furthermore, by releasing our pre-trained models, we aim to reduce the need for redundant, energy-consuming training cycles by the community, thereby reducing the environmental impact.

\end{document}